# Estimates of maize plant density from UAV RGB images using Faster-RCNN detection model: impact of the spatial resolution

K. Velumani,[1,2]* R. Lopez-Lozano,[2]* S. Madec,[3] W. Guo,[4] J. Gillet,[1] A. Comar,[1] F. Baret[2]

[1] Hiphen SAS, 120 rue Jean Dausset, Agroparc, Bâtiment Technicité, 84140 Avignon, France

[2] INRAE, UMR EMMAH, UMT CAPTE, 228 route de l'Aérodrome, Domaine Saint Paul - Site Agroparc CS 40509, 84914 Avignon Cedex 9, France

[3] Arvalis, 228, route de l'Aérodrome – CS 40509, 84914 Avignon Cedex 9, France

[4] International Field Phenomics Research Laboratory, Institute for Sustainable Agro-Ecosystem Services, Graduate School of Agricultural and Life Sciences, The University of Tokyo, Tokyo, Japan

*Corresponding authors. Email : kaaviya.velumani@inrae.fr, raul.lopez-lozano@inrae.fr

## Abstract

Early-stage plant density is an essential trait that determines the fate of a genotype under given environmental conditions and management practices. The use of RGB images taken from UAVs may replace the traditional visual counting in fields with improved throughput, accuracy and access to plant localization. However, high-resolution images are required to detect the small plants present at the early stages. This study explores the impact of image ground sampling distance (GSD) on the performances of maize plant detection at three-to-five leaves stage using Faster-RCNN object detection algorithm. Data collected at high-resolution (GSD»0.3cm) over six contrasted sites were used for model training. Two additional sites with images acquired both at high and low (GSD»0.6cm) resolution were used to evaluate the model performances. Results show that Faster-RCNN achieved very good plant detection and counting (rRMSE=0.08) performances when native high-resolution images are used both for training and validation. Similarly, good performances were observed (rRMSE=0.11) when the model is trained over synthetic low-resolution images obtained by down-sampling the native training high-resolution images, and applied to the synthetic low-resolution validation images. Conversely, poor performances are obtained when the model is trained on a given spatial resolution and applied to another spatial resolution. Training on a mix of high- and low-resolution images allows to get very good performances on the native high-resolution (rRMSE=0.06) and synthetic low-resolution (rRMSE=0.10) images. However, very low performances are still observed over the native low-resolution images (rRMSE=0.48), mainly due to the poor quality of the native low-resolution images. Finally, an advanced super-resolution method based on GAN (generative adversarial network) that introduces additional textural information derived from the native high-resolution images was applied to the native low-resolution validation images. Results show some significant improvement (rRMSE=0.22) compared to bicubic up-sampling approach, while still far below the performances achieved over the native high-resolution images.

## 1. Introduction

Plant density at emergence is an essential trait for crops since it is the first yield component that determines the fate of a genotype under given environmental conditions and management practices [1]–[5]. Competition between plants within the canopy depends on the sowing pattern and its understanding requires reliable observations of the plant localization and density [6]–[9]. An accurate estimation of actual plant density is also necessary to evaluate the seed vigor by linking the emergence rate to the environmental factors [10]–[13].

Maize plant density was measured by visual counting in the field. However, this method is labor-intensive, time consuming and prone to sampling errors. Several higher throughput methods based on optical imagery have been developed in the last twenty years. This was permitted by the technological advances with the increasing availability of small, light and affordable high spatial resolution cameras and autonomous vehicles. Unmanned ground vehicles (UGV) provide access to detailed phenotypic traits [14]–[16] while being generally expensive and associated with throughputs of the order of few hundreds of microplots per hour. Conversely, unmanned aerial vehicles (UAV) are very affordable with higher acquisition throughput than UGVs. When carrying very high-resolution cameras, they can access potentially several traits [17], [18] including plant density [19], [20].

Image interpretation methods used to estimate plant density can be classified into three main categories. The first one is based on machine learning where the plant density measured over a small set of sampling area is related to other canopy level descriptors including vegetation indices derived from RGB and multispectral data [21]–[23]. However, this type of method may lead to significant errors due to the lack of representativeness of the training data set as well as the effect of possible confounding factors including changes in background properties or plant architecture under genetic control. The second category of methods is based on standard computer vision techniques, where the image is first binarized to identify the green objects that are then classified into plants according to the geometrical features defined by the operator (e.g. [24], [25]). The last category of methods is based on deep learning algorithms for automatic object detection [26]–[28]. The main advantage of deep learning methods is their ability to automatically extract low-level features from the images to identify the targeted objects. While deep learning methods appear very promising, their generalization capacity is determined by the volume and diversity of the training dataset [29]. While large collections of images can now be easily acquired, labeling the images used to train the deep models represents a significant effort that is the main limiting factor to build very large training datasets. Few international initiatives have been proposed to share massive labelled datasets that will contribute to maximize the performances of deep learning models [30]–[34], with however questions regarding the consistency of the acquisition conditions and particularly the ground sampling distance (GSD).

The use of UAV images for plant detection at early stages introduces important requirements on image resolution, as deep learning algorithms are sensitive to object scales with the identification of small objects being very challenging [35], [36]. For a given camera, low altitude flights are therefore preferred to get the desired GSD. However, low altitude flights decrease the acquisition throughput because of a reduced camera swath forcing to complete more tracks to cover the same experiment, and requires additionally to slow down the flying speed to reduce motion blur. An optimal altitude should therefore be selected to compromise between the acquisition throughput and the image GSD. Previous studies reporting early-stage maize plant detection from UAVs from deep learning methods did not addressed specifically this important scaling issue [20], [26], [27]. One way to address this scaling issue is to transform the low-resolution images into higher resolution ones using super-resolution techniques. Dai [37] have demonstrated the efficiency of super-resolution techniques to enhance segmentation and edge detection. Later, Fromm [38] and Magoulianitis [39] showed improvements in object detection performances when using the super-resolution methods. The more advanced super-resolution techniques use deep convolutional networks trained over paired high- and low-resolution images [40]–[42]. Since the construction of a real-world paired high- and low-resolution dataset is a complicated task, the high-resolution images are often degraded using a bicubic kernel or less frequently using gaussian noise to constitute the low-resolution images [43]. However more recent studies have shown the drawbacks of the bicubic down-sampling approaches as it smoothens sensor noise and other compression artifacts, thus failing to generalize while applied to real world images [41]. More recent studies propose the use of unsupervised domain translation techniques to generate realistic paired datasets for training the super-resolution networks [44].

We propose here to explore the impact of image GSD on the performances of maize plant detection at stages from three to five leaves using deep learning methods. More specifically, three specific objectives are targeted: (1) to assess the accuracy and robustness of deep learning algorithms for detecting maize plants with high-resolution images used both, in the training and validation datasets; (2) to study the ability of these algorithms to generalize in the resolution domain, i.e. when applied to images with higher and lower resolution compared to the training dataset; and (3) to evaluate the efficiency of data augmentation and preparation techniques in the resolution domain to improve the detection performances. Special emphasis was put here on assessing the contribution of two contrasting methods to up-sample low-resolution images: a simple bicubic up-sampling algorithm, and a more advanced super-resolution model based on GAN (generative adversarial network) that introduces additional textural information. Data collected over several sites across France with UAV flights completed at several altitudes providing a range of GSDs were used.

## 2. Materials and Methods

### 2.1 Study sites

This study was conducted over 8 sites corresponding to different field phenotyping platforms distributed across the west of France and sampled from 2016 to 2019 (Figure 1). The list of sites and their geographic coordinates are given in Table 1. Each platform included different maize microplots with size 20 to 40 square meters. Depending on the experimental design of the platform, the microplots were sown with two to seven rows of maize of different cultivars and row spacing varying from 30 to 110 cm. The sowing dates were always between mid-April and mid-May.

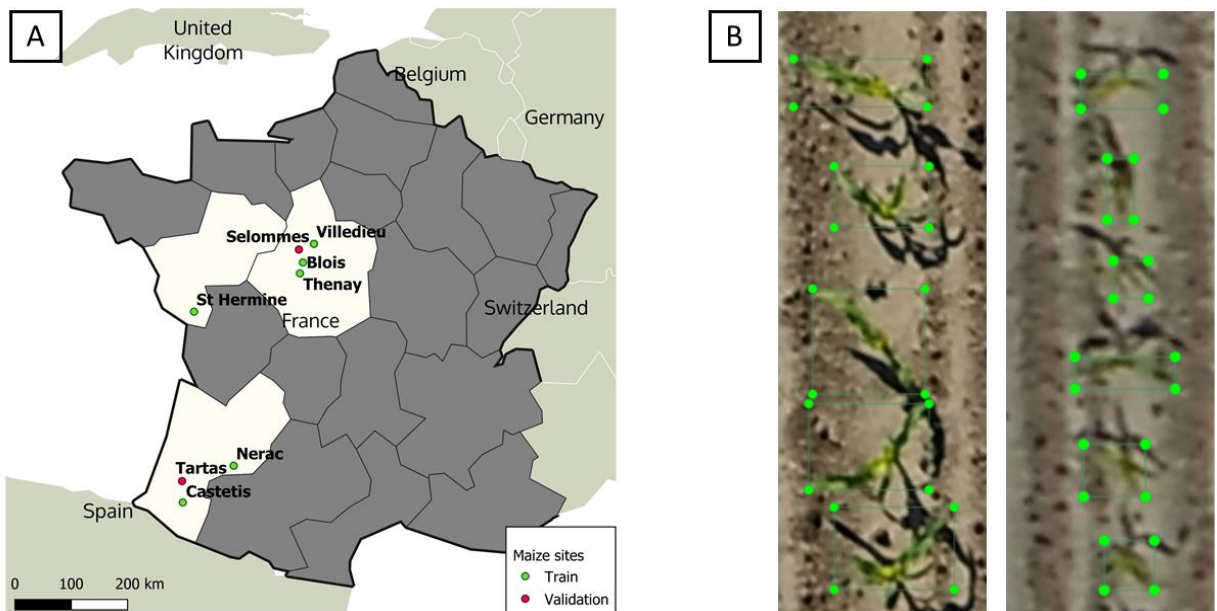

**Figure 1 Location of the study sites with example extracts of the maize microplots acquired from UAV (A)**
A map displaying the location of the eight maize phenotyping platforms located in the west of France used in this study. **(B)** An illustration of the bounding boxes drawn around the maize plants. The examples shown are from the Tartas site (GSD=0.27cm) on the left and Tartas site (GSD=0.63cm) on the right.

**Table 1** The dataset used for the training and validation of the object detection models are listed here. $T^h$, is the training high-resolution dataset, $V^h$ is the validation high-resolution dataset and $V^l$ is the validation low-resolution dataset. * For this site, the microplot extracts were resampled to GSD=0.25 cm before annotating.

| Dataset | # | Site Name | Latitude (°) | Longitude (°) | Acquisition Date | Camera | Flight Altitude (m) | Focal length (mm) | GSD (cm) | Nb. microplots labelled | Nb. plants labelled | Average plant size (pixels) |
|---|---|---|---|---|---|---|---|---|---|---|---|---|
| $T^h$ | 1 | Castetis | 43.46 | -0.71 | 06-06-2019 | FC6540 | 26 | 24 | 0.35 | 20 | 2239 | 939 |
| | 2 | St Hermine | 46.54 | -1.06 | 23-05-2019 | FC6540 | 26 | 35 | 0.33* | 20 | 2674 | 806 |
| | 3 | Nerac | 44.16 | 0.3 | 01-06-2017 | ILCE-6000 | 25 | 30 | 0.32 | 44 | 3338 | 2253 |
| | 4 | Thenay | 47.38 | 1.28 | 18-05-2018 | ILCE-6000 | 22 | 30 | 0.25 | 72 | 7454 | 1505 |
| | 5 | Villedieu | 47.88 | 1.53 | 28-06-2016 | n/a | n/a | n/a | 0.27 | 26 | 2390 | 2159 |
| | 6 | Blois | 47.56 | 1.32 | 18-05-2018 | ILCE-6000 | 25 | 30 | 0.33 | 20 | 1746 | 1419 |
| $V^h$ | 7 | Tartas | 43.80 | -0.79 | 08-06-2019 | FC6540 | 20 | 24 | 0.32 | 22 | 2151 | 1336 |
| | 8 | Selommes | 47.76 | 1.19 | 17-05-2019 | L1D-20c | 16.2 | 10.26 | 0.27 | 14 | 1105 | 891 |
| $V^l$ | 9 | Tartas | 43.80 | -0.79 | 08-06-2019 | FC6540 | 40 | 24 | 0.63 | 24 | 2151 | 437 |
| | 10 | Selommes | 47.76 | 1.19 | 17-05-2019 | L1D-20c | 30 | 10.26 | 0.66 | 14 | 1105 | 156 |

### 2.2 Data acquisition and processing

UAV flights were carried out on the eight sites approximately one month after the sowing date, between mid-May and mid-June (Table 1). Maize plants were in between three to five leaves stage, ensuring that there is almost no overlap among individual plants from near nadir viewing. Three different RGB cameras were used for the data acquisition: Sony Alpha (ILCE-6000) with a focal length of 30 mm, DJI X7 (FC6540) with focal lengths of 24 mm and 30 mm and the default camera with DJI Mavic 2 pro (L1D-20c) with a focal length of 10.26 mm mounted on AltiGator Mikrokopter (Belgium) and DJI Mavic 2 pro (China). To geo-reference the images, ground control points (GCPs) were evenly distributed around the sites and their geographic coordinates were registered using a Real Time Kinematic GPS.

The flights were conducted at an altitude above the ground ranging between 15 and 22 meters, providing a ground sampling distance (GSD) between 0.27 and 0.35 cm (Table 1). For Tartas and Selommes sites, an additional flight was done at a higher altitude on the same day providing a GSD between 0.63 and 0.66 cm.

The flights were planned with a lateral and front overlap of 60/ 80% between individual images. Each dataset was processed using Photoscan Professional (Agisoft LLC, Russia) to align the overlapping images by automatic tie point matching, optimize the aligned camera positions and finally geo-reference the results using the GCPs. The steps followed are similar to the data processing detailed by Madec [15]. Once ortho-rectified, the multiple instances of the microplot present in the overlapping images were extracted using Phenoscript, a software developed within the CAPTE research unit. Phenoscript allows to select, among the individual images available for each microplots, those with full coverage of the microplot, minimum blur and view direction closer to the nadir one. Only these images were used in this study.

### 2.3 Manual labelling of individual plants

From each site, the microplots were labelled with an offline tool, LabelImg [45]: bounding boxes around each maize plant were interactively drawn (Figure 1B). The available sites (Table 1) were divided into three groups: (1) the first group ($T^h$) composed of six sites was used to train the plant detection models. It includes a total of 202 microplots corresponding to 19,841 plants. (2) The second group ($V^h$) corresponding to the Tartas and Selommes with low altitude flights was used to evaluate the model performance at high-resolution. It includes a total of 36 microplots corresponding to 3256 plants. (3) The third group ($V^l$) corresponds to the high-altitude flights in Tartas and Selommes was used to evaluate the model performance at low-resolution. It includes a total of 36 microplots corresponding to 3256 plants. An example of images extracted from the three groups is shown in Figure 2.

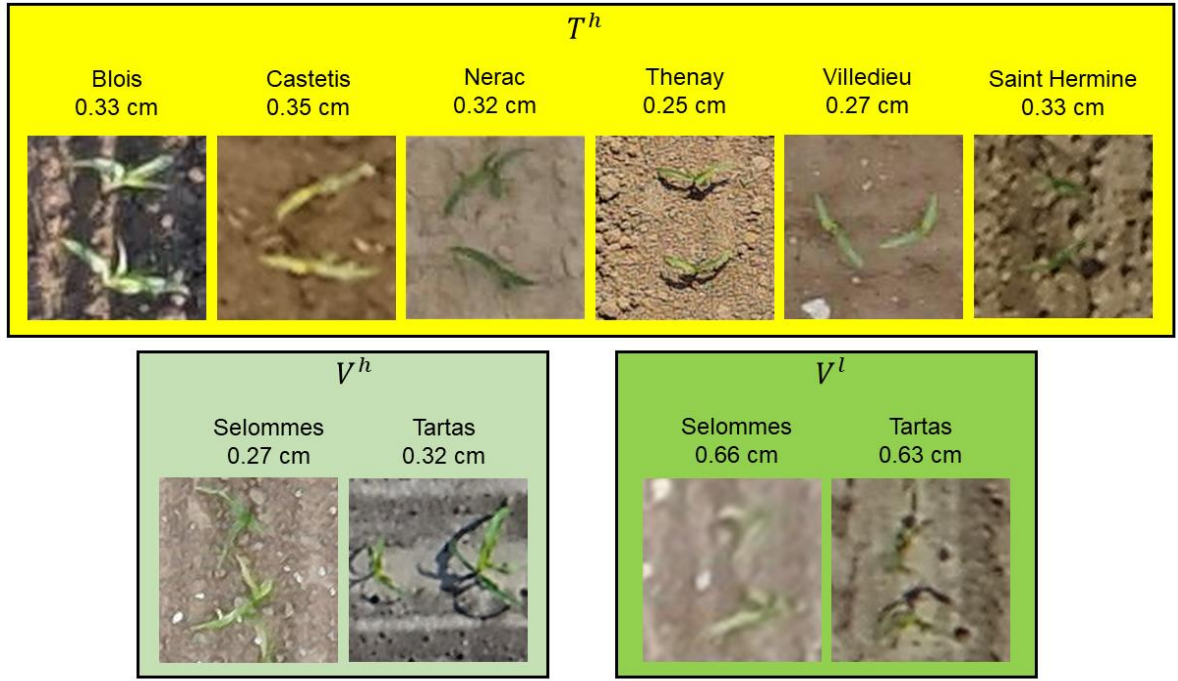

**Figure 2 Examples of maize plants extracted from the in the eight sites used in this study**. The image titles indicate the location of the sites. $T^h$, $V^h$ and $V^l$ are the training high-resolution dataset, validation high-resolution dataset and the validation low-resolution dataset, respectively.

### 2.4 The Faster RCNN object detection model

Faster-RCNN [46], a convolutional neural network designed for object detection was selected to identify maize plants in the image. Besides its wide popularity outside the plant phenotyping community, Faster-RCNN has also been proved to be suitable for various plant and plant-organ detection tasks [47]–[49]. We used the implementation of Faster RCNN in the open-source MMDetection Toolbox [50], written in PyTorch, with pre-trained weights on ImageNet. The Faster-RCNN model with a ResNet50 backbone was trained for 12 epochs with a batch size of 2. The weights were optimized using an SGD optimizer (Stochastic Gradient Descent) with a learning rate of 0.02. For the model training, ten patches of 512 x 512 pixels were randomly extracted from each microplot in the training sites. Standard data augmentation strategies such as rotate, flip, scale, brightness/contrast and jpeg compression were applied.

### 2.5 Experimental Plan

To evaluate the effect of the resolution on the reliability of maize plant detection, we compared Faster RCNN performances over training and validation datasets made of images of high (GSD≈0.30 cm) and low (GSD≈0.60 cm) resolution. Three training datasets built from $T^h$ (Table 1) were considered: (1) the original $T^h$ dataset with around 0.32 cm GSD; (2) A dataset, $T^{h \rightarrow l}_{gm}$ where the images from $T^h$ were down-sampled to 0.64 cm GSD using a gaussian filter and motion blur that mimics the actual low-resolution imagery acquired at higher altitude as described later (section 2.6.1); (3) A dataset, where the original $T^h$ high-resolution dataset was merged with its low-resolution transform, $T_{gm}^{h \to l}$. This $T^h + T_{gm}^{h \to l}$ is expected to provide robustness of the model towards changes in GSD. Note that we did not investigated the training with the native low-resolution images because the labeling of low-resolution images is often difficult because plants are not easy to identify visually and to draw accurately the corresponding bounding box. Further, only two flights were available at the high altitudes (Table 1) that were reserved for the validation. A specific model was trained over each of the three training datasets considered (Table 2) and then evaluated over independent high- and low-resolution validation datasets.

We considered three validation datasets for the high-resolution images: (1) the native high-resolution validation dataset, $V^h$ acquired at low altitude with GSD around 0.30 cm (Table 1). (2) a synthetic high-resolution dataset of GSD around 0.30 cm obtained by up-sampling the native low-resolution dataset, acquired at high altitude, using a bicubic interpolation algorithm as described in section 2.6.2. It will be called $V_{bc}^{l \to h}$. (3) A synthetic high-resolution dataset, $V_{sr}^{l \to h}$, obtained by applying a super-resolution algorithm (see Section 2.6.3) to the native low-resolution dataset $V^l$ and resulting in images with a GSD around 0.30 cm. Finally two low-resolution datasets will be also considered: (1) The native low-resolution validation dataset, $V^l$ (Table 1), with a GSD around 0.60 cm. (2) A synthetic low-resolution dataset, $V_{gm}^{h \to l}$, obtained by applying a Gaussian filter to down-sample (see Section 2.6.1) the original high-resolution dataset, $V^h$, and get a GSD around 0.60 cm.

**Table 2 Description of the training and validation datasets.**

| | Dataset name | Nb. microplots | Nb. plants | Comment |
|---|---|---|---|---|
| Training | $T^h$ | 202 | 19,841 | Native high-resolution training dataset |
| | $T_{gm}^{h \to l}$ | 202 | 19,841 | Down-sampling $T^h$ with gaussian filter and motion blur |
| | $T^h + T_{gm}^{h \to l}$ | 404 | 39,682 | Merging $T^h$ and $T^{h \to l}$ |
| Validation | $V^h$ | 36 | 3256 | Native high-resolution validation dataset |
| | $V_{bc}^{l \to h}$ | 36 | 3256 | Up-sampling $V^l$ with bi-cubic algorithm |
| | $V_{sr}^{l \to h}$ | 36 | 3256 | Up-sampling $V^l$ with Cycle-ESRGAN super-resolution |
| | $V^l$ | 36 | 3256 | Native low-resolution validation dataset |
| | $V_{gm}^{h \to l}$ | 36 | 3256 | Down-sampling $V^h$ with gaussian filter and motion blur |

### 2.6 Methods for image up- and down-sampling
#### 2.6.1 Gaussian filter down-sampling

To create the synthetic low-resolution datasets $T_{gm}^{h \to l}$ and $V_{gm}^{h \to l}$, a Gaussian filter with a sigma=0.63 and a window size=9 followed by a motion blur with a kernel size=3 and angle=45 were applied to down-sample the native high-resolution datasets $T^h$ and $V^h$ by a factor of 2. This solution was preferred to the commonly used bicubic down-sampling method because it provides low-resolution images more similar to the native low-resolution UAV images (Figure 3). This was confirmed by comparing the image variance over the Selommes and Tartas sites where both native high and low-resolution images were available: the variance of the $V_{gm}^{h \to l}$ was closer to that of $V^l$ whereas the bicubic down-sampled dataset had a larger variance corresponding to sharper images. This is consistent with [38] and [51] who used the same method to realistically downsample high-resolution images.

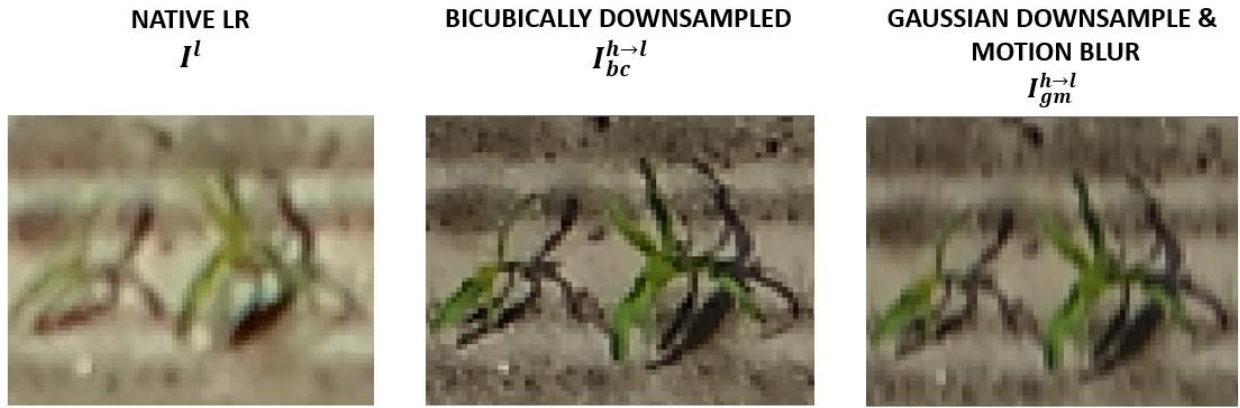

**Figure 3 Visual comparison of the extract of the same plant from the Tartas site between different versions of low-resolution.** Native low-resolution, synthetic low-resolution from bicubic down-sampling, and synthetic low-resolution from Gaussian down-sampling (sigma=0.63, window=9) followed by a motion blur (kernel size=3 and angle=45).

### 2.6.2 Bicubic up-sampling

The bicubic interpolation algorithm was used to generate $V_{bc}^{l \rightarrow h}$ by up-sampling the native low-resolution UAV images, $V^l$. The bicubic interpolation available within Pillow, the Python Imaging Library [52] was used to resample the images.

### 2.6.3 Super-resolution images derived from Cycle-ESRGAN

The super-resolution (SR) is an advanced technique that artificially enhances the textural information while up-sampling images. We used a SR model inspired from [53]. It is a two-stage network composed of a CycleGAN network that generates synthetic paired data and a ESRGAN network capable of image upsampling. The CycleGAN [54] performs unsupervised domain mapping between the native low-resolution and bicubic downsampled low-resolution domains. Thus, for any given input image, CycleGAN is trained to add realistic image noise typical of low-resolution images. The ESRGAN-type super-resolution network [42] up-samples by a factor of two the low-resolution images.

The paired high-resolution and "realistic" low-resolution dataset generated by the CycleGAN was used in the simultaneous training of the ESRGAN-stage of the network. The CycleGAN stage of the network was initially trained for a few epochs following which the two stages (CycleGAN + ESRGAN) were trained together simultaneously. It should be noted that during inference, only the ESRGAN stage of the network would be activated. The training parameters and losses reported by Han [53] were used for the model training. The model weights were initialized over the Div2k dataset [55] and finetuned on the UAV dataset detailed below. The Cycle-ESRGAN network was implemented using Keras [56] deep learning library in Python. The codes will be made available on Github at the following link: https://github.com/kaaviyave/Cycle-ESRGAN

A dedicated training dataset for the super-resolution network was prepared using UAV imagery belonging to the following two domains:
- **Native high-resolution domain**: 2234 microplot extractions from four sites with an average GSD of less than 0.33 cm. Some of the sites belonging to the $T^h$ dataset was used as a part of the training.
- **Native low-resolution domain**: 1713 microplot extractions from three sites with an average GSD of 0.46 cm per site.

None of the validation sites ($V^l$ and $V^h$ in Table 1) were used in the training of the super resolution model. The synthetic downsampled dataset used to train the CycleGAN was prepared by bicubic downsampling the native high-resolution domain by a factor of 2. The images were split into patches of size 256 x 256 pixels for the high-resolution domain and into 128 x 128 pixels for the low-resolution domain.

### 2.7 Evaluation metrics

In this study, the Average Precision (AP), Root Mean Squared Error (RMSE) and Accuracy will be utilized for the evaluation of the Faster-RCNN models for the purpose of maize plant detection and counting.

**AP**: The AP is a frequently used metric for the evaluation of object detection models and can be considered as the area under the precision-recall curve.

$$\text{Precision} = \frac{TP}{TP + FP} \qquad \text{Recall} = \frac{TP}{TP + FN}$$

Where TP is the number of True Positive, FP is the number of False Positive and the FN is the number of False Negative. For the calculation of AP, a predicted bounding box is considered True Positive (TP) if its intersection area over union area (IoU) with the corresponding labelled bounding box is larger than a given threshold. Depending on the objective of the study, different variations exist in the AP metric calculation and the choice of IOU threshold used to qualify a predicted bounding box as TP. After considering several IoU threshold values, we decided to use an IoU threshold of 0.25 to compute AP. This will be later justified. The Python COCO API was used for the calculation of the AP metric [57].

**Accuracy** evaluates the model's performance by calculating the ratio of correctly identified plants to all the predictions made by the model. A predicted bounding box is considered true positive if it has a confidence score of more than 0.5 and an IoU threshold of 0.25. Accuracy is then calculated as:

$$Ac = \frac{TP}{TP + FP + FN}$$

The **relative root mean square error (rRMSE)** between the number of labeled and detected plants across all images belonging to the same dataset:

$$rRMSE = \frac{\sqrt{\frac{\sum_i^n (P_{o,i} - P_{p,i})^2}{n}}}{\overline{P_{o,i}}}$$

where $P_{o,i}$ is the number of plants labeled on image $i$ and $P_{p,i}$ is the number of images predicted by the CNN (confidence score > 0.5 and an IoU > 0.25) and $\overline{P_{o,i}}$ is the average number of labeled plants per image.

## 3. Results and Discussion

### 3.1 Faster RCNN detects plants with high accuracy at high spatial resolution

Very good performances (rRMSE=0.08; Ac=0.88, AP=0.95) are achieved when the model is trained over the high-resolution images ($T^h$) and applied on high-resolution images taken on independent sites ($V^h$). The good performances are explained by the high rate of true positives (Figure 4a). However, the detector performs slightly differently on the two sites used for the validation: in Selommes, an over-detection (false positives, FP) is observed for a small number of plants, when the detector splits a plant into two different objects (Figure 4b). Conversely, in the Tartas site, some under-detection (false negatives, FN) is observed, with a small number of undetected plants (Figure 4).

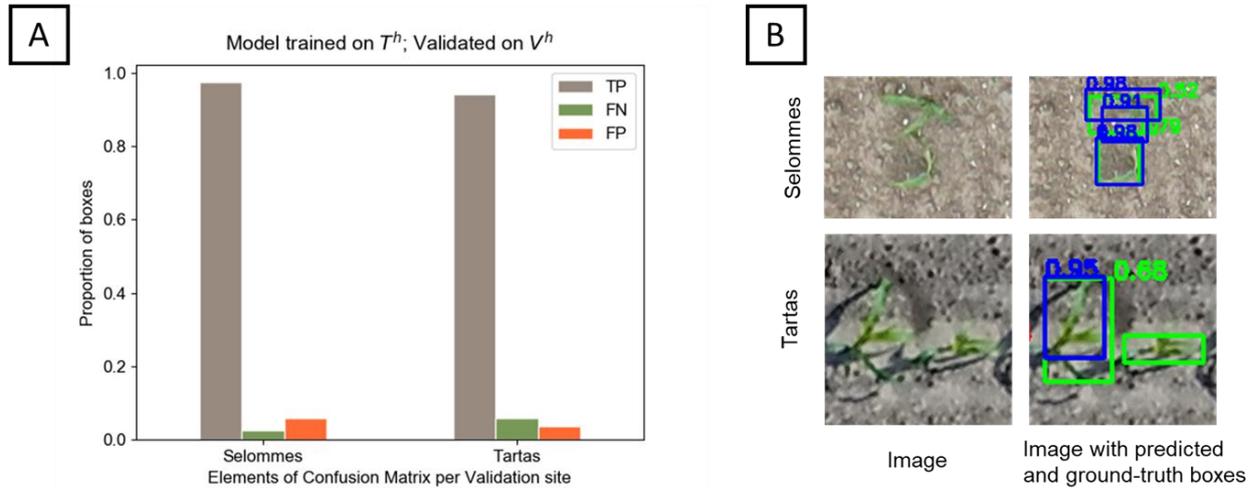

**Figure 4 Results of the model trained on the native HR dataset, $T^h$, and applied to the HR validation dataset, $V^h$. A)** Elements of the confusion matrix - True Positives (TP), False Negatives (FN) and False Positives (FP) for Selommes and Tartas sites. **B)** An example of false positive and false negative observed in the two validation sites. The ground truth bounding boxes are shown in green and the predicted bounding boxes are shown in blue. The green text indicates the IoU of the predicted bounding box with the ground truth and the blue text indicates the confidence score of the predictions.

A detailed analysis of the Precision-Recall curves for the configuration $[T^h, V^h]$ at different IoU (Figure 5) shows a drastic degradation of the detector performances when the IoU is higher than 0.3. This indicates that the model is not accurate when determining the exact dimensions of maize plants. This is partly explained by the difficulty of separating the green from the ground in the shadowed parts of the images. As a consequence, some shaded leaves are excluded from the bounding boxes proposed by the detector and, conversely, some shadowed ground are wrongly included in the bounding boxes proposed (Figure 4b). Further, when a single plant is split into two separate objects by the detector, the resulting bounding boxes are obviously smaller than the corresponding plant (Figure 4b). As a consequence, we proposed to use an IoU threshold of 0.25 to evaluate the model performance to better account for the smaller size of the detected bounding boxes. This contrasts from most object detection applications where an IoU threshold of 0.5 or 0.75 is commonly used to evaluate the performance of the methods [58], [59]. The observed degradation of the model performance for IoU above 0.3 indicates that the method presented provides less accurate localization than in other object detection studies, including both, real world objects and phenotyping applications [49], [60], [61]. An inaccurate estimation of plant dimensions is not critical for those applications assessing germination or emergence rates and uniformity, where plant density is the targeted phenotypic trait. If the focus is to additionally assess the plant size in early developmental stages as well, mask-based RCNN [62], [63] could be used instead. In contrast to algorithms trained on rectangular regions like Faster-RCNN, mask-based algorithms have the potentials to more efficiently manage the shadow projected on the ground by plants, limiting therefore the possible confusion between shaded leaves and ground during the training. However, generating mask annotations is time-consuming, increasing the effort needed to generate a diverse training dataset.

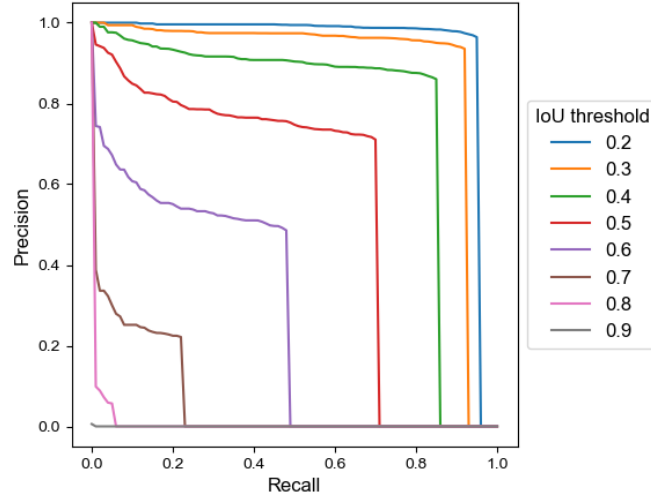

**Figure 5** Precision-Recall curve at different IoU thresholds for the plant detection model trained and applied with high-resolution images ($T^h, V^h$).

These results provide slightly better performances as those reported by David [20] with Ac≈0.8 and rRMSE≈0.1 when using the "out-domain" approach as the one used in this study, i.e. when the training and validation sites are completely independent. They used images with a spatial resolution around 0.3cm as in our study. This is also consistent with the results of Karami [26] who obtained an accuracy of 0.82 with a spatial resolution of around 1cm. They used the anchor-free Few Shot Leaning (FSL) method which identifies and localizes the maize plants by estimating the central position. They claim that their method is little sensitive to object size and thus to the spatial resolution of the images. The accuracy increases up to 0.89 when introducing few images from the validation sites in the training dataset. Kitano [27] proposed a two-step method: they first, segment the images using a CNN-based method and then count the segmented objects. They report an average rRMSE of 0.24 over a test dataset where many factors including image resolution vary (ranging from GSD≈0.3cm to 0.56 cm). They report that their method is sensitive to the size and density of the objects. In the following, we will further investigate the dependency of the performances to image resolution.

### 3.2 The Faster-RCNN model is sensitive to image resolution and apparent plant size

The performances of the model were evaluated when it is trained and validated over images with different resolution. When Faster-RCNN is trained on the high-resolution domain ($T^h$) and applied to a dataset with low-resolution ($V^l$), both AP and Ac decrease almost by 30% (Table 3) compared to the results where the model is trained and applied over high-resolution images. The rate of true positive drops because of the drastic increase of false negatives indicating a high rate of misdetection (Figure 6, $[T^h, V^l]$). This degradation of the detection performances impacts highly the rRMSE that increases up to 0.48. This indicates that the model is sensitive to the resolution of the images. We further investigated if this was linked to the apparent size of the plants and therefore up-sampled the validation low-resolution images with a bicubic interpolation method ($V_{bc}^{l \rightarrow h}$) to get plants with the same size as in the native high-resolution images ($V^h$). Results show that Ac increases from 0.54 to 0.63 and AP from 0.64 to 0.77. However, because of the high imbalance between FN and FP (Figure 6, $[T^h, V_{bc}^{l \rightarrow h}]$), the counting performances remains poor with rRMSE=0.49.

**Table 3 Comparison of the performance of the Faster-RCNN models trained and validated over datasets with different resolution.** The colors indicate the goodness of the performances for the three metrics (green: best; red: worst).

|  |  | Validation | | | | | |
|---|---|---|---|---|---|---|---|
|  |  | High Resolution ($V^h$) | | | Low Resolution ($V^l$) | | |
|  |  | rRMSE | Ac | AP | rRMSE | Ac | AP |
| Training | $T^h$ | 0.08 | 0.88 | 0.95 | 0.48 | 0.54 | 0.64 |
| | $T_{gm}^{h \to l}$ | 0.52 | 0.56 | 0.71 | 0.29 | 0.76 | 0.81 |

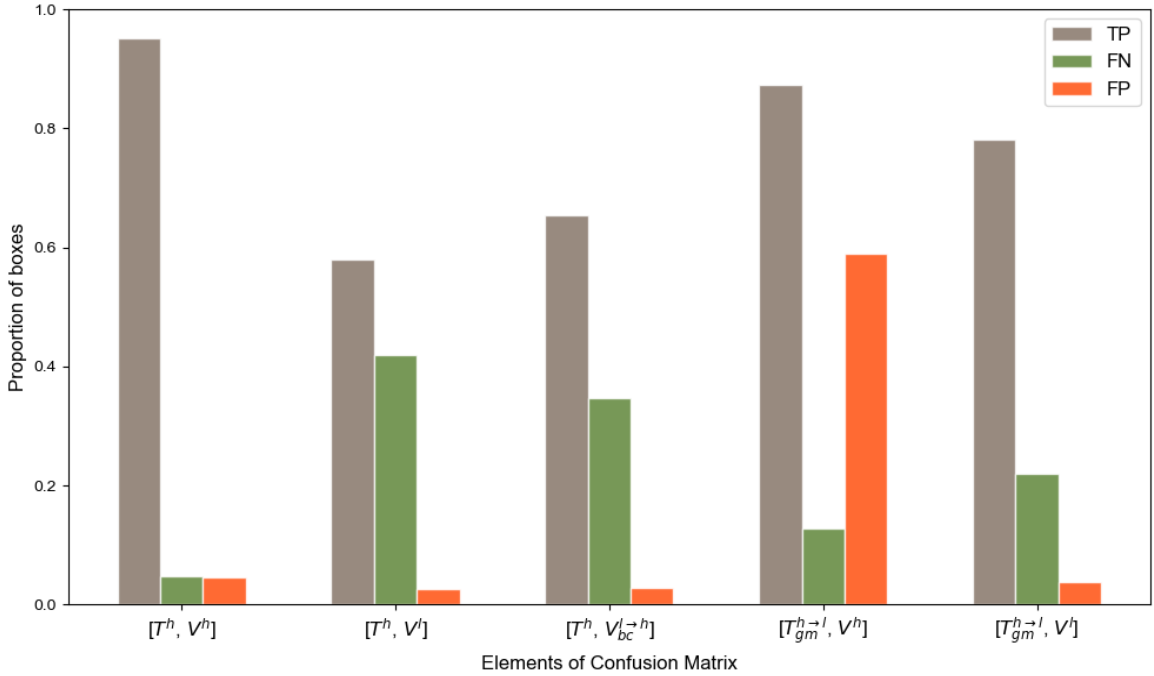

**Figure 6 Results of maize plant detection when trained and evaluated across different resolution domains.** $T^h$ native high-resolution training dataset; $T_{gm}^{h \to l}$ low-resolution training dataset by down-sampling $T^h$ using gaussian motion blur; $V^h$ native high-resolution validation dataset; $V^l$ native low-resolution validation dataset; $V_{bc}^{l \to h}$ high-resolution dataset by up-sampling $V^l$ using bicubic interpolation.

When the model is trained over simulated low-resolution images ($T_{gm}^{h \to l}$), the detection and counting performances evaluated on high-resolution images ($V^h$) also degrades drastically (Table 3). The rate of true positive is relatively high, but the rate of false positive increases drastically (Figure 6 $[T_{gm}^{h \to l}, V^h]$). We observe that the average number of predicted bounding boxes overlapping each labelled box increases linearly with its size (Figure 7). For example, the model identifies on average two plants inside plants larger than 4000 pixels. The imbalance between FN and FP explains the very poor counting performances with rRMSE=0.52 (Table 3). This result confirms the importance to keep consistent the resolution and plant size between the training and the application datasets since Faster-RCNN tends to identify objects that have a similar size to the objects used during the training.

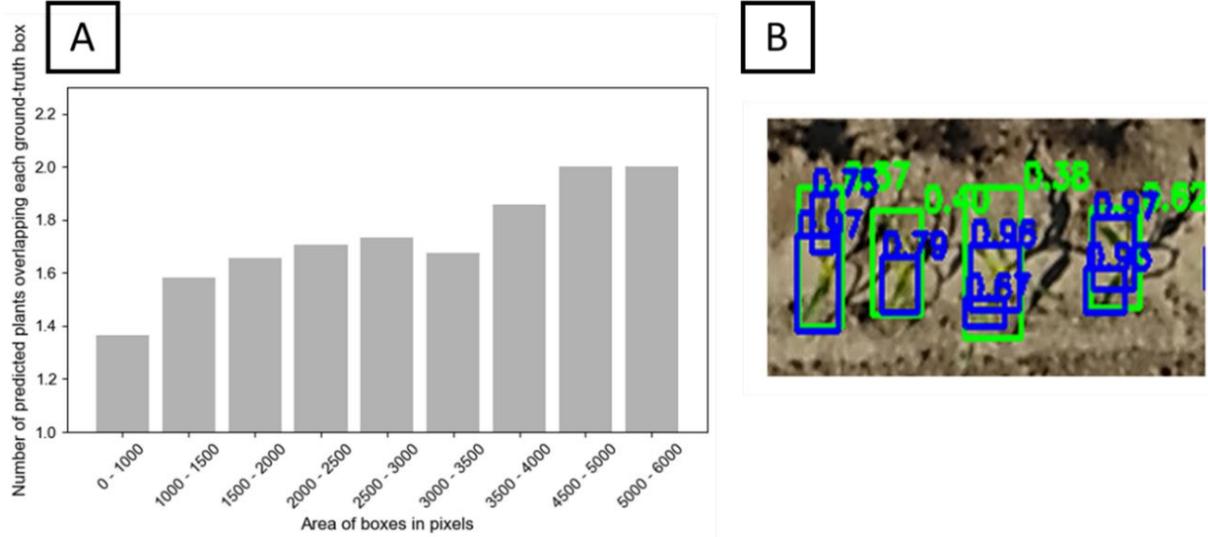

**Figure 7 Effects of the hyper-specialization of Faster-RCNN trained with synthetic low-resolution images $(T_{gm}^{h \to l})$ and applied to a high-resolution datatset ($V^l$):** **A)** Relationship between the size of the ground–truth bounding boxes and the average number of predicted bounding boxes intersecting with them; **B)** Example of over-detection of maize plants due to different object size. The ground truth bounding boxes are shown in green and the predicted boxes are shown in blue. The green text indicates the IoU of the predicted boxes with the ground truth and the blue text indicates the confidence score of the predictions.

We thus evaluated whether data-augmentation may improve the performances on the low-resolution images ($V^l$): the Faster-RCNN model trained on the simulated low-resolution images ($T_{gm}^{h \to l}$) shows improved detection performances as compared to the training over the native high-resolution images (Table 3) with a decrease of the rRMSE down to 0.29 (Table 3). When this model trained with synthetic low-resolution images ($T_{gm}^{h \to l}$) is applied to a dataset downscaled to a similar resolution ($V_{gm}^{h \to l}$), the performances improve dramatically with Ac increasing from 0.56 to 0.89 and AP from 0.71 to 0.90 while the rRMSE drops to 0.10. However, when this model trained with synthetic low-resolution images ($T_{gm}^{h \to l}$) is applied to the native low-resolution images ($V^l$), moderate detection performances are observed which degrades the counting estimates with rRMSE=0.29 (Table 3).

The performances of the model trained over the synthetic low-resolution images ($T_{gm}^{h \to l}$) are quite different when evaluated over the native images ($V^l$) or the synthetic ones ($V_{gm}^{h \to l}$) with the latter yielding results almost comparable to the high-resolution configurations with AP=0.90 (Table 3). This indicates that the low-resolution synthetic images contain enough information to detect accurately the maize plants. Conversely, the native low-resolution image, $V^l$, have probably lost part of the textural information. In addition, the model trained on the synthetic low-resolution images is not able to extract the remaining pertinent plant descriptors from the native low-resolution images. We can observe that the native low-resolution images contain less details as compared to the synthetic ones (Figure 8): some plants are almost not visible in the $V^l$ images, as the textural information vanishes and even the color of maize leaves cannot be clearly distinguished from the soil background. This explains why the model was not able to detect the plants, even when it is trained with the synthetic low-resolution images ($T_{gm}^{h \to l}$). Contrary to vectors that operate at an almost constant height like ground vehicles [16], [64]–[66] or fixed cameras [67]–[70], camera settings (aperture, focus and integration time) in UAVs need to be adapted to the flight conditions, especially flight altitude, to maximize image quality. Further, the jpg recording format of the images may also significantly impact image quality.

Recording the images in raw format would thus improve the detection capability at the expense of increased data volume and sometimes image acquisition frequency.

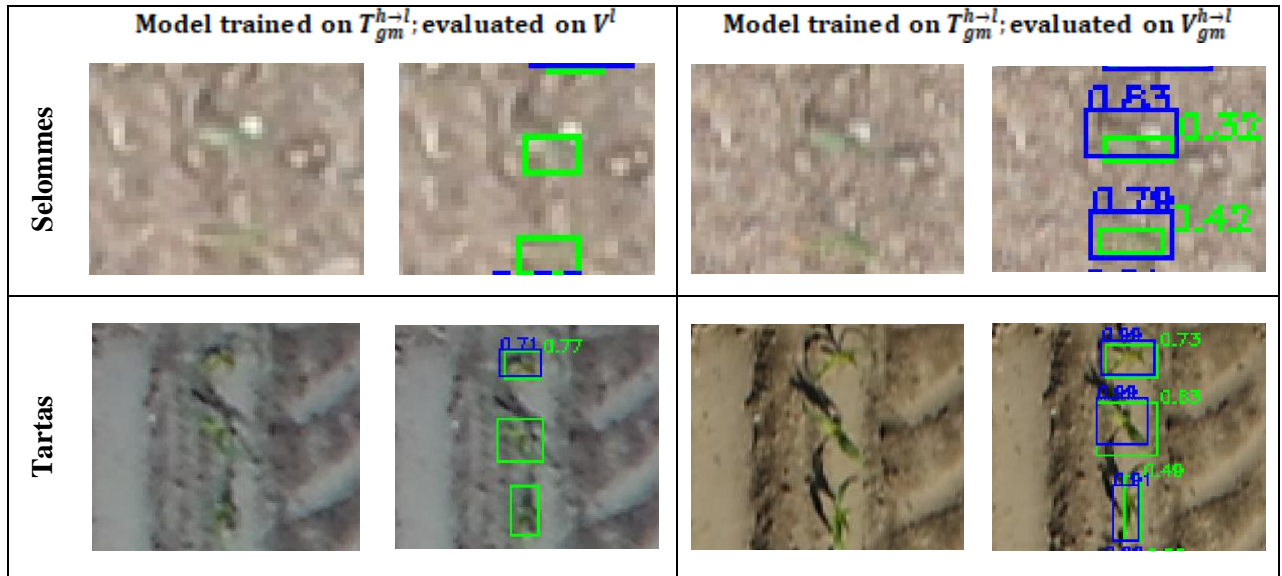

**Figure 8 An example showing the same plants extracted from the exact same locations in two versions of the validation dataset: native LR ($V^l$) and the synthetic LR obtained from gaussian downsampling ($V_{gm}^{h \to l}$).** The first and third column show the raw images while the second and fourth column shows the detector predictions. The ground truth bounding boxes are shown in green and the predicted bounding boxes are shown in blue. The green text indicates the IoU of the predicted box with the ground truth and the blue text indicates the confidence score of the predictions.

### 3.3 Data Augmentation makes the model more resistant to changes in image resolution

We finally investigated whether mixing high and low-resolution images in the training dataset would make the model more resistant to changes in the image resolution. Results show that merging native high-resolution with synthetic low-resolution images ($T^h + T_{gm}^{h \to l}$) provides (Table 4) performances similar to those observed when the model is trained only over high ($T^h$) or synthetic low ($T_{gm}^{h \to l}$) and validated on the same resolution ($V^h$ or $V_{gm}^{h \to l}$) (Table 3). This proves that data augmentation could be a very efficient way to deal with images having different resolution. Further, this model trained on augmented data ($T^h + T_{gm}^{h \to l}$) (Table 4) surprisingly beats the performances of the model trained only on the high-resolution images ($T^h$) as displayed in Table 3. This is probably a side effect of the increase of the size of the training dataset (Table 2). Nevertheless, when validating on the native low-resolution images ($V^l$) (Table 4) the performances are relatively poor as compared to the model trained only on the synthetic low-resolution images ($T_{gm}^{h \to l}$). This is explained by the lower quality of the native low-resolution images as already described in the previous section.

**Table 4 Comparison of the performance of the Faster-RCNN models trained over the augmented data ($T^h + T_{gm}^{h \to l}$) and validated over datasets with different resolution.** The colors indicate the goodness of the performances for the three metrics (green: best; red: worst)**.**

|  | rRMSE | Ac | AP |
|---|---|---|---|
| $V^h$ | 0.06 | 0.91 | 0.96 |
| $V_{gm}^{h \to l}$ | 0.11 | 0.92 | 0.91 |
| $V^l$ | 0.48 | 0.64 | 0.72 |

## 3.4 Up-sampling with the super-resolution method improves the performances of plant detection on the native low-resolution images

If the training is difficult with the native low-resolution images because plants are visually difficult to identify and label, the training should be done over low-resolution images derived from the high-resolution images using a more realistic up-sampling method than the standard bicubic interpolation one. Alternatively, the training could be done using the high-resolution images and the low-resolution dataset may be up-sampled to a synthetic high-resolution domain using bicubic interpolation or super-resolution techniques.

Results show that the super-resolution technique improved plant detection very significantly as compared to the native low-resolution ($V^l$) and bi-cubic up-sampled ($V^{l \to h}_{bc}$) images (Table 5). This impacts positively the counting performances while not reaching the performances obtained with the high-resolution images ($V^h$). The super-resolution reduces drastically the under-detection of maize plants particularly on the Tartas site (Figure 9) where, as mentioned in Section 3.2, these native low-resolution images have lower textural information and green fraction per plant.

**Table 5 Comparison of the performance of the Faster-RCNN models trained over high-resolution images and applied to the native low-resolution images ($V^l$), the synthetic high-resolution images that is up-sampled/transformed using either bi-cubic ($V^{l \to h}_{bc}$) or super-resolution ($V^{l \to h}_{sr}$) techniques.** The colors indicate the goodness of the performances for the three metrics (green: best; red: worst).

|  | rRMSE | Ac | AP |
|---|---|---|---|
| $V^l$ | 0.58 | 0.54 | 0.64 |
| $V^{l \to h}_{bc}$ | 0.43 | 0.63 | 0.77 |
| $V^{l \to h}_{sr}$ | 0.22 | 0.80 | 0.85 |

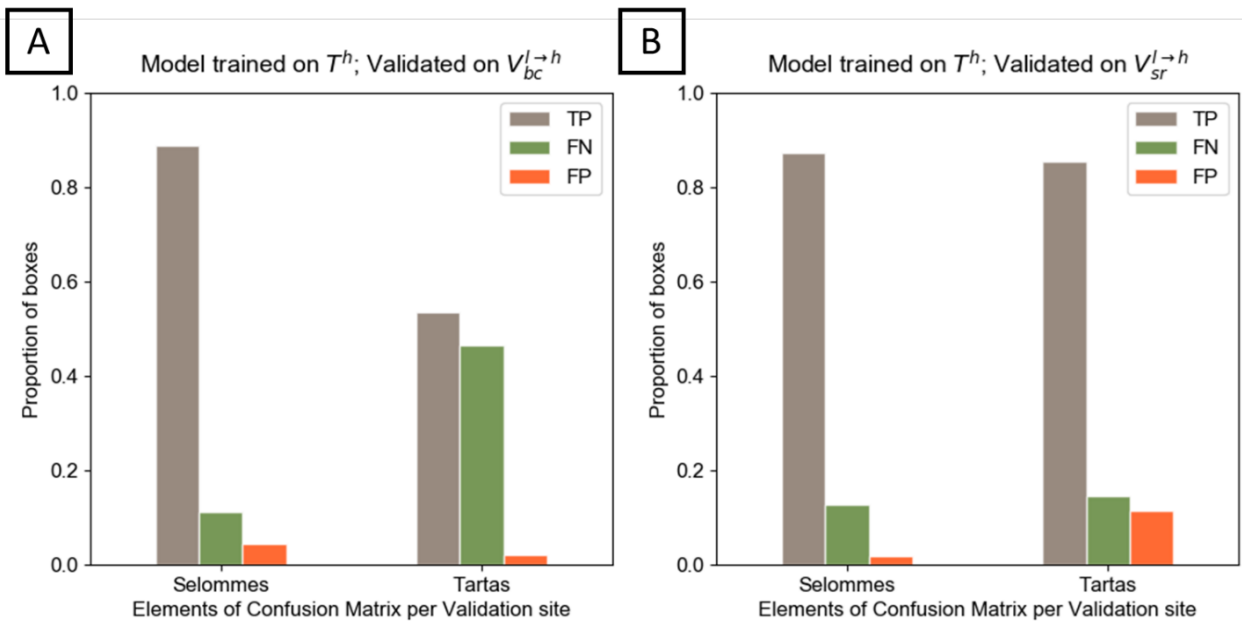

**Figure 9 A comparison between the performance of the Faster-RCNN model trained on the HR dataset, $T^h$, and applied to the synthetic high-resolution datasets. A)** Model trained on $T^h$ and evaluated on the synthetic

HR dataset $V_{bc}^{l \to h}$, from the bicubic up-sampling; **B)** Model trained on $T^h$ applied to the synthetic HR dataset $V_{sr}^{l \to h}$, from the super-resolution technique.

The super resolution approach enhances the features used to identify maize plants, with colors and edges more pronounced than in the corresponding native LR images (Figure 10). Maize plants are visually easier to recognize in the super-resolved images as compared to both the native low-resolution and the bicubically up-sampled images.

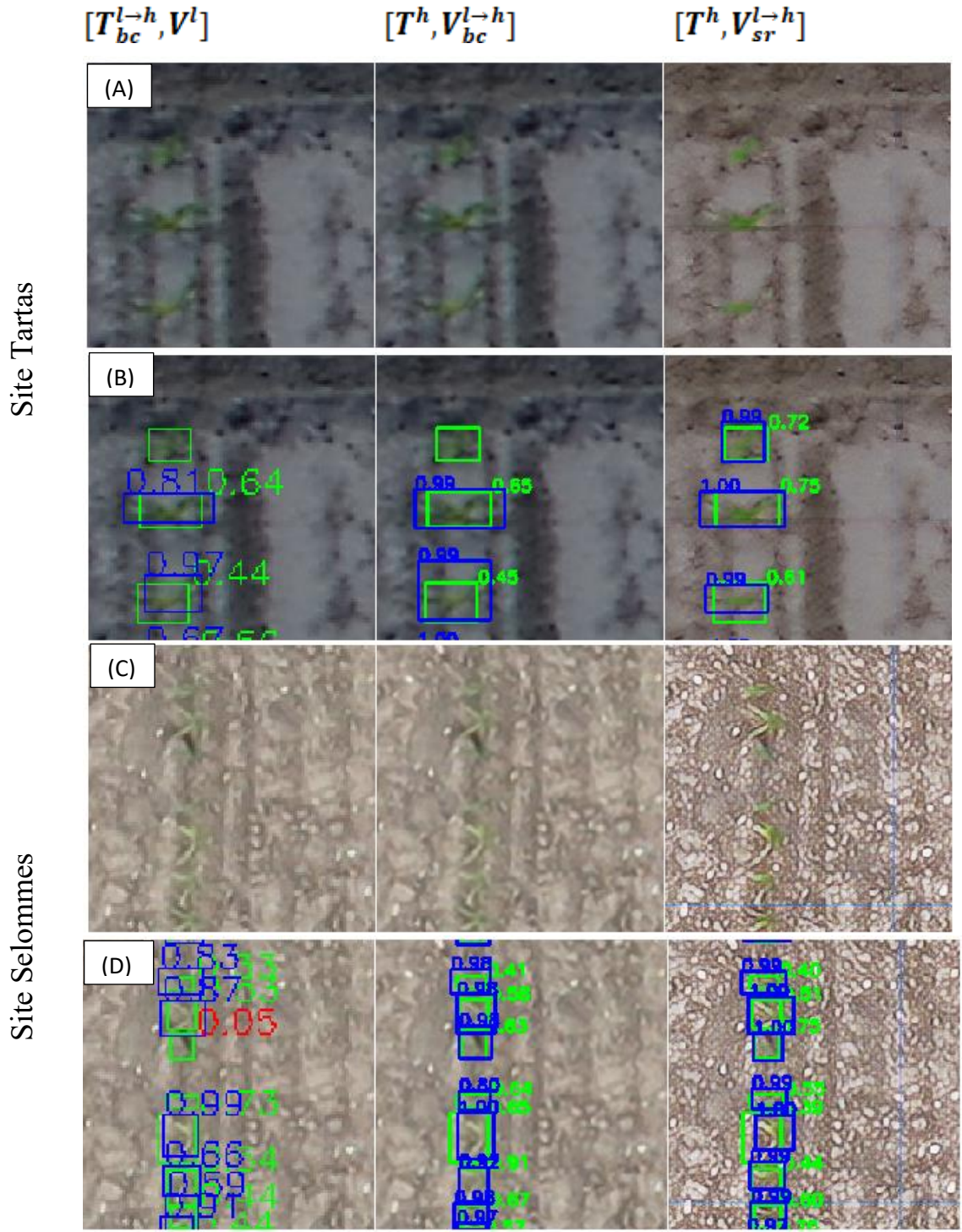

**Figure 10 Illustration of the performance of the Faster-RCNN model on the synthetic high-resolution and native low-resolution datasets.** **(A) and (C):** Images belonging to three datasets - native low-resolution $V^l$, bicubically up-sampled $V_{bc}^{l \to h}$ and finally the up-sampling by super-resolution technique $V_{sr}^{l \to h}$. **(B) and (D):** The

results predicted by the model trained on $T_{gm}^{h \rightarrow l}$ applied to $V^l$ (first column) and the model trained on $T^h$ applied to the synthetic high-resolution datasets $V_{bc}^{l \rightarrow h}$ and $V_{sr}^{l \rightarrow h}$ (second and third column). The ground truth bounding boxes are shown in green and the predicted bounding boxes are shown in blue. The green text indicates the IoU of the predicted bounding box with the ground truth and the blue text indicates the confidence score of the predictions.

Nevertheless, although easier to interpret, the images generated by super-resolution do not appear natural with some exaggerated textural features of the soil background (Figure 10 c, d). In few cases, super-resolution images show new features – e.g. coloring some pixels in green– in leaf-shaped shadows or tractor tracks in the background leading to an increase in the proportion of false positives in certain microplots of the Tartas site (Figure 9b). Training the super-resolution model with a larger dataset might help the generator network to limit those artifacts. Alternatively, some studies [39], [71], [72] have proposed to integrate the training of the super-resolution model with the training of the Faster-RCNN. The use of a combined detection loss would provide additional information on the location of the plants, thus forcing the super-resolution network to differentiate between plants and background while up-sampling the images.

## 4. Conclusion

We evaluated the performances of automatic maize plant detection from UAV images using deep learning methods. Our results show that the Faster-RCNN model achieved very good plant detection and counting (rRMSE=0.08) performances when high-resolution images (GSD≈0.3cm) are used both for training and validation. However, when this model is applied to the low-resolution images acquired at higher altitudes, the detection and counting performances degrade drastically with rRMSE=0.48. We demonstrated that this was mostly due to the hyper-specialization of Faster-RCNN that is expecting plants of similar size as in the training dataset. The sensitivity of the detection method to the object size is a critical issue for plant phenotyping applications, where datasets can be generated from different platforms (UAVs, ground vehicles, portable imaging systems, etc.) each one of them providing images within at a specific ground resolution. Concurrently, it would be optimal to share labeled images to get a wide training dataset. Data augmentation techniques where high and low-resolution images populate the training dataset were proved to be efficient and provides performances similar to the ones achieved when the model is trained and validated over the same image resolution. However, the native low-resolution images acquired from the UAV have significant low quality that prevents accurate plant detection. In some cases, the images are difficult to visually interpret which poses a problem both for their labeling and for the detector to localize plants due to the lack of pertinent information. These low-quality images were characterized by a loss of image texture that could come from camera intrinsic performances, inadequate settings and the jpg recording format. It is thus recommended to pay a great attention to the camera choice, settings and recording format when the UAV is flying at altitudes that provides resolution coarser than 0.3 cm for maize plant counting.

Finally, we evaluated a super-resolution Cycle-ESRGAN based method to partially overcome the problem of sub-optimal image quality. The super-resolution method significantly improved the results on the native low-resolution dataset compared to the classic bicubic up-sampling strategies. However, the performances when applied to the native low-resolution images were moderate and far poorer than those obtained with the native high-resolution images with simulated super-resolved images showing sometimes artifacts. A future direction to reduce the artifacts of such super-resolution algorithms can be to integrate the GAN training along with the training of the plant detection network. Another direction would be to introduce some labeled low-resolution images in the training dataset to possibly integrate their features in model.

**Acknowledgments**

**General**: We would like to thank the CAPTE team for contributing to the construction of the labelled dataset used in this study.

**Author contributions:** KV, RL, SM and FB designed the study. KV implemented the super-resolution pipeline and conducted the analysis. KV and RL contributed extensively to the writing of the article. RL and FB supervised the study. AC and WG participated in discussions and in writing the manuscript. All authors read, revised and approved the final manuscript.

**Funding:** We received support from ANRT for the CIFRE grant of KV, co-funded by Hiphen.

**Competing interests:** The authors declare that there is no conflict of interest regarding the publication of this article.